\documentclass[journal]{IEEEtran}
\ifCLASSINFOpdf
\else
\fi
\usepackage{graphicx}
\usepackage{subfigure}
\usepackage{amsmath}
\usepackage[noadjust]{cite}
\usepackage{xcolor}
\usepackage{float}
\usepackage{cite}
\usepackage{enumerate}
\usepackage{cases}
\usepackage{multirow}
\usepackage{bm}
\usepackage{booktabs}
\usepackage{array}
\usepackage[ruled]{algorithm2e}
\usepackage{makecell}

\begin{document}
\title{Semi-Heterogeneous Three-Way Joint Embedding Network for Sketch-Based Image Retrieval}

\author{Jianjun~Lei,~\IEEEmembership{Senior Member,~IEEE,}
        Yuxin~Song,
        Bo~Peng,
        Zhanyu~Ma,~\IEEEmembership{Senior Member,~IEEE,}\\
        Ling~Shao,~\IEEEmembership{Senior Member,~IEEE,}
        ~and~Yi-Zhe Song
\thanks{This work was supported in part by the Natural Science Foundation of Tianjin ( No.18ZXZNGX00110, 18JCJQJC45800 ), and National Natural  Science Foundation of China ( No.61931014, 61922015, 61722112 ). Copyright 20xx IEEE. Personal use of this material is permitted. However, permission to use this material for any other purposes must be obtained from the IEEE by sending an email to pubs-permissions@ieee.org. (Corresponding author: Bo Peng)\emph{}}
\thanks{J. Lei, Y. Song, and B. Peng are with the School of Electrical and Information Engineering, Tianjin University, Tianjin 300072, China (e-mail: jjlei@tju.edu.cn; songyuxin@tju.edu.cn; bpeng@tju.edu.cn).}
\thanks{Z. Ma is with the Pattern Recognition and Intelligent System Laboratory, Beijing University of Posts and Telecommunications, Beijing 100876, China (e-mail: mazhanyu@bupt.edu.cn).}
\thanks{L. Shao is with the Inception Institute of Artificial Intelligence, Abu Dhabi, United Arab Emirates (e-mail: ling.shao@ieee.org).}
\thanks{Y.-Z. Song is with the SketchX Lab, Centre for Vision, Speech and Signal Processing, University of Surrey, Guildford, Surrey GU2 7XH, U.K. (e-mail: y.song@surrey.ac.uk).}
\thanks{Digital Object Identifier}}

\markboth{}%
{Shell \MakeLowercase{\textit{et al.}}: Bare Demo of IEEEtran.cls for IEEE Journals}

\maketitle

\begin{abstract}
Sketch-based image retrieval (SBIR) is a challenging task due to the large cross-domain gap between sketches and natural images. How to align abstract sketches and natural images into a common high-level semantic space remains a key problem in SBIR. In this paper, we propose a novel semi-heterogeneous three-way joint embedding network (Semi3-Net), which integrates three branches (a sketch branch, a natural image branch, and an edgemap branch) to learn more discriminative cross-domain feature representations for the SBIR task. The key insight lies with how we cultivate the mutual and subtle relationships amongst the sketches, natural images, and edgemaps. A semi-heterogeneous feature mapping is designed to extract bottom features from each domain, where the sketch and edgemap branches are shared while the natural image branch is heterogeneous to the other branches. In addition, a joint semantic embedding is introduced to embed the features from different domains into a common high-level semantic space, where all of the three branches are shared. To further capture informative features common to both natural images and the corresponding edgemaps, a co-attention model is introduced to conduct common channel-wise feature recalibration between different domains. A hybrid-loss mechanism is designed to align the three branches, where an alignment loss and a sketch-edgemap contrastive loss are presented to encourage the network to learn invariant cross-domain representations. Experimental results on two widely used category-level datasets (Sketchy and TU-Berlin Extension) demonstrate that the proposed method outperforms state-of-the-art methods.
\end{abstract}


\IEEEpeerreviewmaketitle

\section{Introduction}
\IEEEPARstart{S}{INCE} the number of digital images on the Internet has increased dramatically in recent years, content-based image retrieval technology has become a hot topic in computer vision community \cite{1}-\cite{4}. With the popularization of touch screen devices, sketch-based image retrieval (SBIR) has attracted extensive attention and achieved remarkable performance \cite{5}-\cite{8}. Given a hand-drawn sketch, the SBIR task aims to retrieve the natural target images from the image database. However, compared with the natural target images, which are full of color and texture information, sketches only contain simple black and white pixels \cite{9}, \cite{10}. Therefore, hand-drawn sketches and natural images belong to two heterogeneous data domains, and aligning these two domains into a common feature space remains the most challenging problem in SBIR.

Traditional SBIR methods describe sketches and natural images using hand-crafted features \cite{11}-\cite{19}. Edgemaps of natural images are usually first extracted as sketch approximations. Then, hand-designed features, such as HOG \cite{20}, SIFT \cite{21}, and Shape Context \cite{22}, are exploited to describe both the sketches and edgemaps.  Finally, the K-Nearest Neighbor (KNN) ranking process is utilized to evaluate the similarity between the sketches and natural images to obtain the final retrieval results. However, as mentioned above, hand-drawn sketches and natural images belong to two heterogeneous data domains. It is difficult to design a common type of feature applicable to two different data domains. Besides, sketches are usually drawn by non-professionals, making them full of intra-class variations \cite{23}-\cite{25}. Most hand-crafted features have difficulties in dealing with these intra-class variations and ambiguities of hand-drawn sketches, which also negatively impacts the performance of SBIR.

In recent years, convolutional neural networks (CNNs) have been widely used across fields \cite{26}-\cite{29}, such as person re-identification, object detection, and video recommendation. In contrast to traditional hand-crafted methods, CNNs can automatically aggregate shallow features learned from the bottom convolutional layers. Inspired by the learning ability of CNNs, several Siamese networks and Triplet networks have been proposed for the SBIR task \cite{30}-\cite{33}. Most of these methods encode a sketch-image or a sketch-edgemap pair, and learn the similarity between the input pair using a contrastive loss or triplet loss. However, there are still several difficulties and challenges to be solved in these methods. 1) The different characteristics of sketches and natural images make the SBIR task challenging. Generally, a sketch only contains the object to be retrieved, meaning the sketches tend to have relatively clean backgrounds. In addition, since sketches are usually drawn by non-professionals, the shapes of objects in sketches are usually deformed and relatively abstract. For natural images, although the objects are not usually significantly deformed, natural images taken by cameras usually have complex backgrounds. Therefore, creating a network that can learn more discriminative features for both sketches and natural images remains a challenge. 2) Since sketches and natural images are from two different data domains, there exists a significant domain gap between the features of these two. Most deep SBIR methods that adopt a contrastive loss or a triplet loss to learn the cross-domain similarity are not effetive enough to cope with the intrinsic domain gap. Therefore, finding a way to eliminate or reduce the cross-domain gap and embedding features from different domains into a common high-level semantic space is critical for SBIR. 3) More importantly, most existing methods achieve SBIR by exploring the matching relationship between either sketch-edgemap pairs or sketch-image pairs. However, the methods using sketch-edgemap pairs ignore the discriminative features contained in natural images, while the methods using sketch-image pairs ignore the auxiliary role of edgemaps. Enabling full use of the joint relationships among sketches, natural images, and edgemaps provides a novel way to solve the cross-domain learning problem.

To address the above issues, a novel semi-heterogeneous three-way joint embedding network (Semi3-Net) is proposed in this paper to mitigate the domain gap and align sketches, natural images, and edgemaps into a high-level semantic space. The key insight behind our design is how we enforce mutual cooperation amongst the three branches. We importantly recognize that when measured in terms of visual abstraction, sketches and edgemaps are more closely linked compared with sketches and natural images. The sketches are highly abstract and iconic representations of natural images, and edgemaps are reduced versions of natural images, where detailed appearance information such as texture and color are removed. However, compared with edgemaps, natural images contain more discriminative features for SBIR. Motivated by this insight, we purposefully design a semi-heterogeneous joint embedding network, where a semi-heterogeneous weight-sharing setting among the three branches is adopted in the feature mapping part, while a three-branch all-sharing setting is conducted in the joint semantic embedding part. This design essentially promotes edgemaps to act as a ``bridge'' to help narrow the domain gap between the natural images and sketches. Fig. \ref{figure1} offers a visualization of the proposed Semi3-Net architecture. More specifically, the semi-heterogeneous feature mapping part is designed to extract the bottom features for each domain, where a co-attention model is introduced to learn informative features common between different domains. Meanwhile, the joint semantic embedding part is proposed to embed the features from different domains into a common high-level semantic space. In addition, a hybrid-loss mechanism is proposed to achieve a more discriminative embedding, where an alignment loss and a sketch-edgemap contrastive loss are introduced to encourage the network to learn invariant cross-domain representations. The main contributions of this paper are summarized as follows.

1) A novel semi-heterogeneous three-way joint embedding network is proposed, in which the semi-heterogeneous feature mapping and the joint semantic embedding are designed to learn joint feature representations for sketches, natural images, and edgemaps.

2) To capture informative features common between natural images and the corresponding edgemaps, a co-attention model is developed between the natural image and edgemap branches.

3) A hybrid-loss mechanism is designed to mitigate the domain gap, where an alignment loss and a sketch-edgemap contrastive loss are presented to encourage the network to learn invariant cross-domain representations.

4) Experiments on two widely-used datasets, Sketchy and TU-Berlin Extension, demonstrate that the proposed method outperforms state-of-the-art methods.

The rest of the paper is organized as follows. Section II reviews the related works. Section III introduces the proposed method in detail. The experimental results and analysis are presented in Section IV. Finally, the conclusion is drawn in Section V.

\section{Related Work}

\subsection{Traditional SBIR Methods}
Traditional SBIR methods usually utilize edge extraction methods to extract edgemaps from natural images first. Then, hand-crafted features are used as descriptors for both sketches and edgemaps. Finally, a KNN ranking process within a Bag-of-Words (BoW) framework is usually utilized to rank the candidate natural images for each sketch. For instance, Hu \emph{et al.} \cite{11} incorporated the Gradient Field HOG (GF-HOG) into a BoW scheme for SBIR, and obtained promising performance. Saavedra \emph{et al.} \cite{12} introduced Soft-Histogram of Edge Local Orientations (SHELO) as the descriptors for sketches and edgemaps extracted from natural images, which effectively improves the retrieval accuracy. In \cite{13}, a novel method for describing hand-drawn sketches was proposed by detecting learned keyshapes (LKS). Xu \emph{et al.} \cite{14} proposed an academic coupled dictionary learning method to address the cross-domain learning problem in SBIR. Qian \emph{et al.} \cite{15} introduced re-ranking and relevance feedback schemes to find more similar natural images based on initial retrieval results, thus improving the retrieval performance.

\subsection{Deep SBIR Methods}
\begin{figure*}[!t]
\centering
\includegraphics[width=1\linewidth]{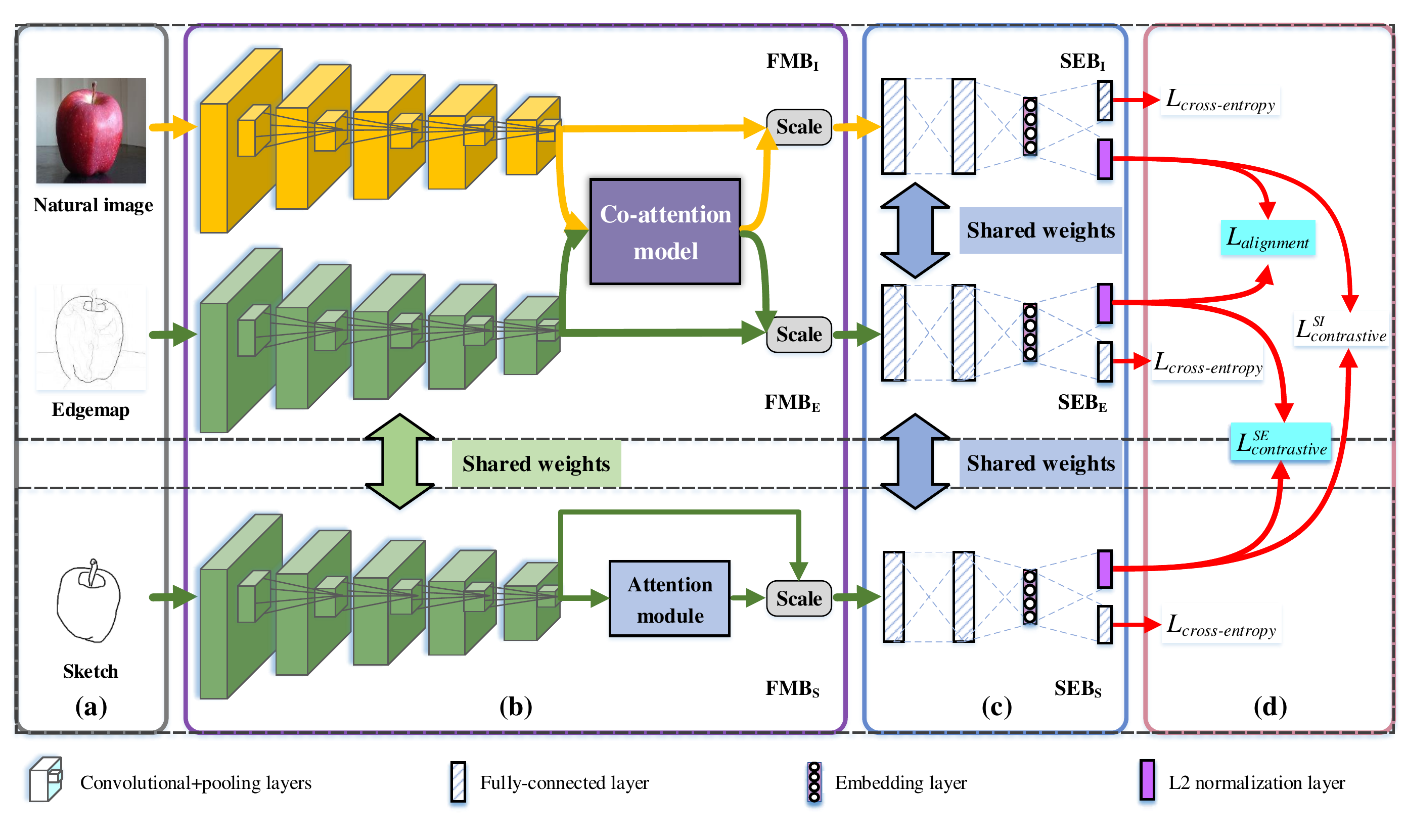}
\setlength{\abovecaptionskip}{-0.75cm}
\caption{ Illustration of the proposed Semi3-Net. (a) Three-way inputs. (b) Semi-heterogeneous feature mapping. (c) Joint semantic embedding. (d) Hybrid-loss mechanism. Blocks with the same color represent their weights are shared.}
\label{figure1}
\end{figure*}
Recently, many frameworks based on CNNs have been proposed to address the challenges in SBIR \cite{33}-\cite{39}. Aiming to learn the cross-domain similarity between the sketch and natural image domains, several Siamese networks have been proposed to improve the retrieval performance. Qi \emph{et al.} \cite{33} introduced a novel Siamese CNN architecture for SBIR, which learns the features of sketches and edgemaps by jointly tuning two CNNs. Liu \emph{et al.} \cite{34} proposed a Siamese-AlexNet based on two AlexNet \cite{40} branches to learn the cross-domain similarity and mitigate the domain gap. Wang \emph{et al.} \cite{36} proposed a Siamese network to learn the similarity between the input sketches and edgemaps of 3D models, which is originally designed for sketch-based 3D shape retrieval. Meanwhile, several Triplet architectures have also been proposed, which include the sketch branch, the positive natural image branch, and the negative natural image branch. In these methods, a ranking loss function is utilized to constrain the feature distance between a sketch and a positive natural image to be smaller than the one between the sketch and a negative natural image. Sangkloy \emph{et al.} \cite{38} learned a cross-domain mapping through a pre-training strategy to embed natural images and sketches in the same semantic space, and achieved superior retrieval performance. Recently, deep hashing methods \cite{54}, \cite{55} have been exploited for retrieval task and have achieved significant improvement on retrieval performance. Liu \emph{et. al.} \cite{34} integrated a deep architecture into the hashing framework to capture the cross-domain similarities and speed up the SBIR process. Zhang \emph{et al.} \cite{39} proposed a Generative Domain-migration Hashing (GDH) approach, which uses a generative model to migrate sketches to their indistinguishable natural image counterparts and achieves the best-performing results on two SBIR datasets.

\subsection{Attention Models}
Attention models have recently been successfully applied to various deep learning tasks, such as natural language processing (NLP) \cite{41}, fine-grained image recognition \cite{42}, \cite{43}, video moment retrieval \cite{44}, and visual question answering (VQA) \cite{45}. In the field of image and video processing, the two most commonly used attention models are soft-attention models \cite{46} and hard-attention models \cite{47}. Soft-attention models assign different weights to different regions or channels in an image or a video, by learning an attention mask. In contrast, hard attention models only indicate one region at a time, using reinforcement learning. For instance, Hu \emph{et al.} \cite{48} proposed a channel attention model to recalibrate the weights of different channels, which effectively enhances the discriminative power of features and achieves promising classification performance. Gao \emph{et al.} \cite{56} proposed a novel aLSTMs framework for video captioning, which integrates the attention mechanism and LSTM to capture salient structures for video. In \cite{57}, a hierarchical LSTM with adaptive attention approach was proposed for image and video captioning, and achieved the state-of-the-art performances on both tasks. Besides, Li \emph{et al.} \cite{47} proposed a harmonious attention network for person re-identification, where soft attention is used to learn important pixels for fine-gained information matching, and hard attention is applied to search latent discriminative regions. Song \emph{et al.} \cite{35} proposed a soft attention spatial method for fine-grained SBIR to capture more discriminative fine-grained features. This model reweights the different spatial regions of the feature map by learning a weight mask for each branch of the triplet network. However, although the attention mechanisms above have strong abilities for feature learning, they generally learn discriminative features using only the input itself. For the SBIR task, we are more concerned about learning discriminative cross-domain features for retrieval. In other words, the common features of different domains should be considered simultaneously. To address this, a co-attention model is exploited in this paper for the SBIR task, to focus on capturing informative features common between natural images and the corresponding edgemaps, and further mitigate the cross-domain gap.

\section{Semi-Heterogeneous Three-Way Joint Embedding Network}

\subsection{Semi-Heterogeneous Feature Mapping} As shown in Fig. \ref{figure1} (b), the semi-heterogeneous feature mapping part consists of the natural image branch ${\rm{FM}}{{\rm{B}}_{\rm{I}}}$, the edgemap branch ${\rm{FM}}{{\rm{B}}_{\rm{E}}}$, and the sketch branch ${\rm{FM}}{{\rm{B}}_{\rm{S}}}$. Each branch includes a series of convolutional and pooling layers, which aim to learn the bottom features for each domain. As mentioned above, sketches and edgemaps have similar characteristics, and both lack detailed appearance information such as texture and color, thus the weights of ${\rm{FM}}{{\rm{B}}_{\rm{S}}}$ and ${\rm{FM}}{{\rm{B}}_{\rm{E}}}$ are shared. Besides, since the amount of the sketch training data is much smaller than natural image training data, sharing weights between the sketch and edgemap branches can partly alleviate problems that arise from the lack of sketch training data.
\begin{figure}[tbp]
\centering
\includegraphics[width=1\linewidth]{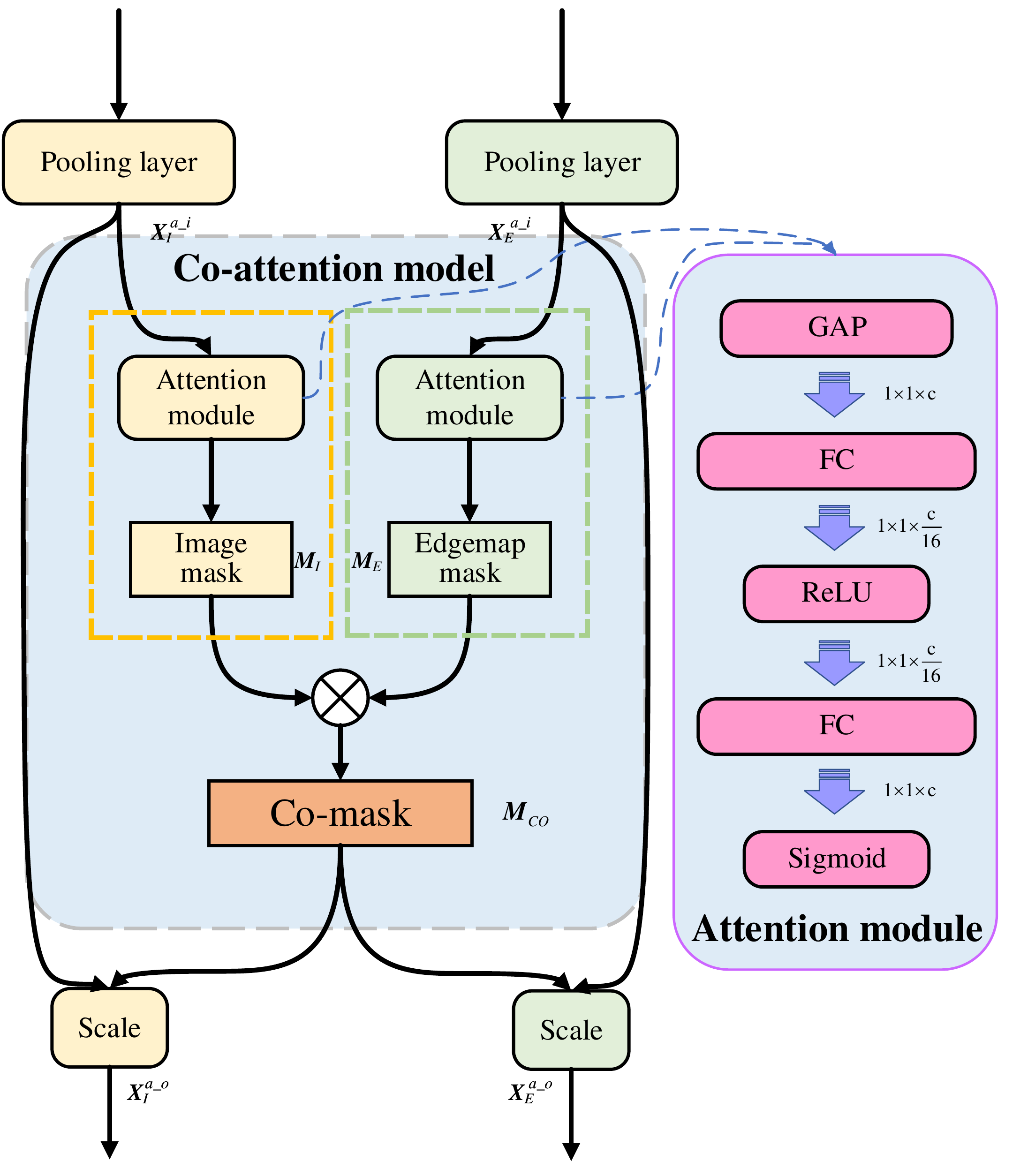}
\setlength{\abovecaptionskip}{-0.3cm}
\caption{ Structure of the co-attention model. Yellow blocks represent the layers that belong to the natural image branch, while green blocks represent the layers that belong to the edgemap branch.}
\label{figure2}
\end{figure}

Meanwhile, since there exist obvious features representation differences between natural images and sketches (edgemaps), the bottom convolutional layers of the natural image branch should be learned separately. Accordingly, the natural image branch ${\rm{FM}}{{\rm{B}}_{\rm{I}}}$ does not share weights with the other two branches in the feature mapping part. Additionally, with the aim of learning the informative features common between different domains, a co-attention model is introduced to associate the ${\rm{FM}}{{\rm{B}}_{\rm{I}}}$ and ${\rm{FM}}{{\rm{B}}_{\rm{E}}}$ branches. By applying the proposed co-attention model, the network is able to focus on the discriminative features common to both natural images and the corresponding edgemaps, and discard information that is not important for the retrieval task. For the sketch branch ${\rm{FM}}{{\rm{B}}_{\rm{S}}}$, a channel-wise attention module is also introduced to learn more discriminative features. The co-attention model will be discussed in Section III-B in detail.

Particularly, the semi-heterogeneous in the proposed architecture refers to the three-branch semi-heterogeneous weight sharing strategy. The weights of the sketch and edgemap branches are shared while the natural image branch is independent of the others in the semi-heterogeneous feature mapping part. The three branches are integrated into the semi-heterogeneous feature mapping architecture and interact with each other, which ensures that not only are the bottom features of each domain preserved, but that the features from different domains are also prealigned through the co-attention model.

\subsection{Joint Semantic Embedding} In the joint semantic embedding part, as shown in Fig. \ref{figure1} (c), the natural image branch ${\rm{SE}}{{\rm{B}}_{\rm{I}}}$, the edgemap branch ${\rm{SE}}{{\rm{B}}_{\rm{E}}}$, and the sketch branch ${\rm{SE}}{{\rm{B}}_{\rm{S}}}$ are developed to embed the features from different domains into a common high-level semantic space. Each branch of the joint semantic embedding part includes several fully-connected (FC) layers. An extra embedding layer, followed by an L2 normalization layer, is also introduced in each branch. As previously stated, the bottom features of different branches are learned respectively, by the semi-heterogeneous feature mapping. In order to achieve feature alignment between the natural images, edgemaps, and sketches in a common high-level semantic space, the weights of ${\rm{SE}}{{\rm{B}}_{\rm{I}}}$, ${\rm{SE}}{{\rm{B}}_{\rm{E}}}$, and ${\rm{SE}}{{\rm{B}}_{\rm{S}}}$ are completely shared in the joint semantic embedding part. Based on the features learned in the common high-level semantic space, a hybrid-loss mechanism is proposed to learn the invariant cross-domain representations, and achieve a more discriminative embedding.

\subsection{Co-attention Model}
To capture the informative features common between natural images and the corresponding edgemaps, a co-attention model is proposed between the natural image and edgemap branches. As previously stated, edgemaps are reduced versions of natural images, where detailed appearance information are removed. Therefore, a natural image and its corresponding edgemap are spatially aligned, which can be made full use of to narrow the gap between the natural images and edgemaps. Additionally, since the weights of the sketch and edgemap branches are shared, the introduction of the co-attention model actually enforces mutual and subtle cooperation amongst three branches. Specifically, before the bottom features of the three branches are fed into the joint semantic embedding part, the co-attention model prealigns different domains by conducting common feature recalibration.
As shown in Fig. \ref{figure2}, the proposed co-attention model includes two attention modules and a co-mask learning module. The co-attention model takes the output feature maps of the last pooling layers in the natural image and edgemap branches as the inputs,  and learns a common mask which is used to re-weight each channel of the  feature maps in both two branches.

In the attention module, a channel-wise soft attention mechanism is adopted to capture discriminative features of each input. Specifically, the attention module on the right of Fig. \ref{figure2} consists of a global average pooling (GAP) layer, two FC layers, a ReLU layer, and a sigmoid layer. Let ${\bm{X}}_I^{a\_i} \in {R^{h \times {\rm{w}} \times {\rm{c}}}}$ and ${\bm{X}}_E^{a\_i} \in {R^{h \times {\rm{w}} \times {\rm{c}}}}$ denote the inputs of the attention module for the natural image branch and edgemap branch, respectively, where h, w, and c represent the height, width, and channel dimensions of the feature maps. Through the GAP layer, the feature descriptors for natural image and edgemap branches are obtained by aggregating the global spatial information of ${\bm{X}}_I^{a\_i}$ and ${\bm{X}}_E^{a\_i}$, which can be formulated as:
\begin{equation}\begin{array}{c}
{\bm{X}}_I^{gap} = \frac{1}{{h{\rm{ \times w}}}}\sum\limits_{u = 1}^h {\sum\limits_{v = 1}^w {{\bm{X}}_I^{a\_i}(u,v)} }
\end{array}\label{1}\end{equation}
\begin{equation}\begin{array}{c}
{\bm{X}}_E^{gap} = \frac{1}{{h{\rm{ \times w}}}}\sum\limits_{u = 1}^h {\sum\limits_{v = 1}^w {{\bm{X}}_E^{a\_i}(u,v)} }
\end{array}\label{2}\end{equation}

Based on ${\bm{X}}_I^{gap}$ and ${\bm{X}}_E^{gap}$, two FC layers and a ReLU layer are applied to model the channel-wise dependencies, and the attention maps in both two domains are obtained. By deploying a sigmoid layer, each channel of the attention map is normalized to ${[0,1]}$. For different domains, the final learned image attention mask ${{\bm{M}}_I} \in {R^{{\rm{1}} \times {\rm{1}} \times {\rm{c}}}}$ and edgemap attention mask ${{\bm{M}}_E} \in {R^{1 \times 1 \times c}}$ are respectively formulated as:
\begin{equation}\begin{array}{c}
{{\bm{M}}_I} = {sigmoid}({\bm{W}}_{_I}^2 \times {\mathop{ReLU}\nolimits} ({\bm{W}}_I^1 \times {\bm{X}}_I^{gap}))
\end{array}\label{3}\end{equation}
\begin{equation}\begin{array}{c}
{{\bm{M}}_E} = {sigmoid}({\bm{W}}_{_E}^2 \times {\mathop{ReLU}\nolimits} ({\bm{W}}_{_E}^1 \times {\bm{X}}_E^{gap}))
\end{array}\label{4}\end{equation}
where ${\bm{W}}_I^1$ and ${\bm{W}}_E^1$ denote the weights of the first FC layer, and ${\bm{W}}_I^2$ and ${\bm{W}}_E^2$ denote the weights of the second FC layer.

In SBIR, the key challenge is to capture the discriminative information common to different domains and then align the different domains into a common high-level semantic space. Therefore, unlike most existing works that use the obtained attention mask directly to reweight the channel responses, the proposed co-attention model tries to capture the channel-wise dependencies common between different domains by learning a co-mask. Based on the learned image attention mask and edgemap attention mask, the co-mask ${{\bm{M}}_{CO}}$ is defined as:
\begin{equation}\begin{array}{c}
{{\bm{M}}_{CO}} = {{\bm{M}}_I} \otimes {{\bm{M}}_E}
\end{array}\label{5}\end{equation}
where $\otimes$ denotes the element-wise product. Elements in ${{\bm{M}}_{CO}} \in {R^{{\rm{1}} \times {\rm{1}} \times {\rm{c}}}}$ represent the joint weights of the corresponding channels in ${\bm{X}}_I^{a\_i}$ and ${\bm{X}}_E^{a\_i}$.

Afterwards, the output feature maps ${\bm{X}}_I^{a\_o} \in {R^{h \times {\rm{w}} \times {\rm{c}}}}$ and ${\bm{X}}_E^{a\_o} \in {R^{h \times {\rm{w}} \times {\rm{c}}}}$ for the image and edgemap branches are calculated respectively by rescaling the input feature maps ${\bm{X}}_I^{a\_i}$ and ${\bm{X}}_E^{a\_i}$ with the obtained channel-wise co-mask:
\begin{equation}\begin{array}{c}
{\bm{X}}_{_I}^{a\_o} = {f_{scale}}({{\bm{M}}_{CO}},{\bm{X}}_{_I}^{a\_i})
\end{array}\label{6}\end{equation}
\begin{equation}\begin{array}{c}
{\bm{X}}_E^{a\_o} = {f_{scale}}({{\bm{M}}_{CO}},{\bm{X}}_{_E}^{a\_i})
\end{array}\label{7}\end{equation}
where ${f_{scale}}( \cdot )$ denotes the channel-wise multiplication between the co-mask and the input feature maps.

The proposed co-attention model not only considers the channel-wise feature responses of each domain by introducing the attention mechanism, but also captures the common channel-wise relationship between different domains, simultaneously. Specifically, by introducing the co-attention model between the natural image and edgemap branches, the common informative features from different domains are highlighted, and the cross-domain gap is effectively reduced.

\subsection{Hybrid-Loss Mechanism}
SBIR aims to obtain a retrieval ranking by learning the similarity between a query sketch and the natural images in a dataset. The distance between the positive sketch-image pairs should be smaller than the distance between the negative sketch-image pairs. In the hybrid-loss mechanism, the alignment loss and the sketch-edgemap contrastive loss are presented to learn the invariant cross-domain representations, and mitigate the domain gap. Besides, we also introduce the cross-entropy loss \cite{38} and the sketch-image contrastive loss \cite{30}, which are two typical losses in SBIR. The four types of loss functions are complementary to each other, thus improving the separability between the different sample pairs.

To be clear, the feature maps produced by the L2 normalization layers for the three branches are denoted as ${f_{{{\bm{\theta }}_{I}}}}(I)$, ${f_{{{\bm{\theta }}_{E}}}}(E)$, and ${f_{{{\bm{\theta }}_{S}}}}(S)$, where ${f_{\bm{\theta }}}( \cdot )$ denotes the mapping function learned by the network branch, and ${{\bm{\theta }}_I}$, ${{\bm{\theta }}_E}$, and ${{\bm{\theta }}_S}$ denote the weights of the natural image, edgemap, and sketch branches, respectively.

\textbf{1) Alignment loss.} To learn the invariant cross-domain representations and align different domains into a high-level semantic space, a novel alignment loss is proposed between the natural image branch and the edgemap branch. Although the image and the corresponding edgemap come from different data domains, they should have similar high-level semantics after the processing of joint semantic embedding. Motivated by this, aiming to minimize the feature distance between an image and its corresponding edgemap in the high-level semantic space, the proposed alignment loss function is defined as:
\begin{equation}\begin{array}{c}
{L_{alignment}}(I,E) = ||{f_{{{\bm{\theta }}_I}}}(I) - {f_{{{\bm{\theta }}_E}}}(E)||_2^2
\end{array}\label{8}\end{equation}

By introducing the alignment loss, the cross-domain invariant representations between a natural image and its corresponding edgemap are captured for the SBIR task. In other words, the proposed alignment loss provides a novel way of dealing with the domain gap by constructing a correlation between the natural image and its corresponding edgemap. It potentially encourages the network to learn the discriminative features common to both natural image and sketch domains, and successfully aligns different domains into a common high-level semantic space.

\textbf{2) Sketch-edgemap contrastive loss.} Considering the one-to-one correlation between an image and its corresponding edgemap, the sketch-edgemap contrastive loss $L_{contrastive}^{SE}(S,E,{l_{sim}})$ between the sketch and edgemap branches is proposed to further constrain the matching relationship between the sketch-image pair as follows.
\begin{equation}\begin{array}{c}
\begin{array}{c}
L_{contrastive}^{SE}(S,E,{l_{sim}}) = {l_{sim}}d({f_{{{\bm{\theta }}_S}}}(S),{f_{{{\bm{\theta }}_E}}}({E^ + }))\\
 \qquad + (1 - {l_{sim}})\max (0,{m_1} - d({f_{{{\bm{\theta }}_S}}}(S),{f_{{{\bm{\theta }}_E}}}({E^ - })))
\end{array}
\end{array}\label{9}\end{equation}
where ${l_{sim}}$ denotes the similarity label, with 1 indicating a positive sketch-edgemap pair and 0 being a negative sketch-edgemap pair, ${E^ + }$ and ${E^ - }$ denote the edgemap corresponding to the positive and negative natural image, respectively, $d( \cdot )$ denotes Euclidean distance, and ${m_1}$ denotes the margin. The sketch-edgemap contrastive loss aims to measure the similarity between input pairs from the sketch and edgemap branches, thus further aligning different domains into the high-level semantic space.

\textbf{3) Cross-entropy loss.} In order to learn the discriminative features from each domain, the cross-entropy losses [38] for the three branches are introduced. For each branch of the proposed network, a softmax cross-entropy loss ${L_{cross - entropy}}({\bm{p}},{\bm{y}})$ is exploited, which is formulated as:
\begin{equation}\begin{array}{c}
\begin{aligned}
{L_{cross - entropy}}({\bm{p}},{\bm{y}}) &=  - \sum\limits_k^K {{y_k}\log {p_k}}
\\& =- \sum\limits_k^K {{y_k}\log (\frac{{\exp ({z_k})}}{{\sum\nolimits_{j = 1}^K {\exp ({z_j})} }}} )
\end{aligned}
\end{array}\label{10}\end{equation}
where ${\bm{p}} = ({p_1},...{p_K})$ denotes the discrete probability distribution of one data sample over $K$ categories, ${\bm{y}} = ({y_1},...{y_K})$ denotes the typical one-hot label corresponding to each category, and ${\bm{z}} = ({z_1},...{z_K})$ denotes the feature vector produced by the last FC layer. In the proposed Semi3-Net, the cross-entropy loss forces the network to extract the discriminative features for each domain.

\textbf{4) Sketch-image contrastive loss.} Intuitively, in the SBIR task, the positive sketch-image pair should be close together, while the negative sketch-image pair should be far apart. Given a sketch $S$ and a natural image $I$, the sketch-image contrastive loss \cite{30} can be represented as:
\begin{equation}\begin{array}{c}
\begin{array}{c}
L_{contrastive}^{SI}(S,I,{l_{sim}}) = {l_{sim}}d({f_{{{\bm{\theta }}_S}}}(S),{f_{{{\bm{\theta }}_I}}}({I^ + }))\\
\qquad + (1 - {l_{sim}})\max (0,{m_2} - d({f_{{{\bm{\theta }}_S}}}(S),{f_{{{\bm{\theta }}_I}}}({I^ - })))
\end{array}
\end{array}\label{11}\end{equation}
where ${I^ + }$ and ${I^ - }$ denote the positive and negative natural images, respectively, and ${m_2}$ denotes the margin. By utilizing the sketch-image contrastive loss, the cross-domain similarity between sketches and natural images is effectively measured.

Finally, the alignment loss in Eq. \eqref{8}, the sketch-edgemap contrastive loss in Eq. \eqref{9}, the cross-entropy loss in Eq. \eqref{10}, and the sketch-image contrastive loss in Eq. (11) are combined, thus the overall loss function $L(S,I,E,{{\bm{p}}_D},{{\bm{y}}_D},{l_{sim}})$ is derived as:
\begin{equation}\begin{array}{c}
\begin{aligned}
L(S,I,E,{{\bm{p}}_D},{{\bm{y}}_D},{l_{sim}}) &= \sum\limits_{D = I,E,S} {{L_{cross - entropy}}({{\bm{p}}_D},{{\bm{y}}_D})} \\
&{\rm{ + }}\alpha L_{contrastive}^{SI}(S,I,{l_{sim}})\\
&{\rm{ + }}\beta {L_{alignment}}(I,E)\\
&{\rm{ + }}\gamma L_{contrastive}^{SE}(S,E,{l_{sim}})
\end{aligned}
\end{array}\label{12}\end{equation}
where $\alpha $, $\beta $, and $\gamma $ denote the hyper-parameters that control the trade-off among different types of losses.

The proposed hybrid-loss mechanism constructs the correlation among sketches, edgemaps, and natural images, providing a novel way to deal with the domain gap by learning the invariant cross-domain representations. By adopting the hybrid-loss mechanism, the proposed network is able to learn more discriminative feature representations and effectively align the sketches, natural images, and edgemaps into a common feature space, thus improving the retrieval accuracy.
\subsection{Implementation Details and Training Procedure}
In the proposed Semi3-Net, each branch is constructed based on VGG19 \cite{49}, which consists of sixteen convolutional layers, five pooling layers and three FC layers. In terms of network architecture, the convolutional layers and pooling layers in the semi-heterogeneous feature mapping part, and the first two FC layers in the joint semantic embedding part are the same as the ones in VGG19 in structure. Specifically, the extra embedding layer in each branch is a 256-dimension fully-connected layer, which aims to embed the different domains into the common semantic feature space. Meanwhile, the size of the last FC layer is modified to the number of categories in the SBIR datasets. The entire training process includes the pre-training stage on each individual branch and the training stage on the proposed Semi3-Net.

In the pre-training stage, each individual branch, including the convolutional layers and pooling layers in the semi-heterogeneous feature mapping part, and the FC layers in the joint semantic embedding part, is trained independently. Specifically, the cross-entropy loss is adopted to pre-train each branch using the corresponding source data in the training dataset. Without learning the common embedding, the pre-training stage aims to learn the weights appropriate for sketch, natural image and edgemap recognition, respectively.

In the training stage, the weights of the three branches are learned jointly, and the cross-domain representations are obtained by training the whole Semi3-Net. The overall loss function in Eq. \eqref{12} is utilized in the training stage. As for the sketch-edgemap and the sketch-image contrastive losses illustrated above, the sketch-image and sketch-edgemap pairs for training need to be generated. To this end, for each sketch in the training dataset, we randomly select a natural image (edgemap) from the same category to form the positive pair and a natural image (edgemap) from the other categories to form the negative pair. In the training process, the ratio of positive and negative sample pairs is set to 1:1, and the positive and negative pairs are randomly selected following this rule, for each training batch.

\section{Experiments}

\subsection{Experimental Settings}
In this paper, two category-level SBIR benchmarks, Sketchy \cite{38} and TU-Berlin Extension \cite{34}, are adopted to evaluate the proposed Semi3-Net. Sketchy consists of 75,471 sketches and 12,500 natural images, from 125 categories with 100 objects per category. In our experiments, we utilize the extended Sketchy dataset \cite{34} with 73,002 natural images in total, which adds an extra 60,502 natural images collected from ImageNet \cite{50}. TU-Berlin Extension \cite{34} consists of 204,489 natural images and 20k free-hand drawn sketches from 250 categories, with 80 sketches per category. The natural images in these two datasets are all realistic with complex backgrounds and large variations, thus bringing great challenges to the SBIR task. Importantly, for fair comparison against state-of-the-art methods, the same training-testing splits used for the existing methods are adopted in our experiments. For Sketchy and TU-berlin Extension, 50 and 10 sketches, respectively, of each category are utilized as the testing queries, while the rest are used for training.

All experiments are performed under the simulation environments of GeForce GTX 1080 Ti GPU and Intel i7-8700K processor @3.70 GHz. The training process of the proposed Semi3-Net is implemented using SGD on Caffe \cite{51} with a batch size of 32. The initial learning rate is set to 2e-4, and the weight decay and the momentum are set to 5e-4 and 0.9, respectively. For both datasets, the balance parameters $\alpha $, $\beta $, and $\gamma $ are set to 10, 100, and 10, respectively, as they consistently yield promising result. The margins ${m_1}$ and ${m_2}$  for the two contrastive losses are both set to 0.3.

In the proposed method, the Gb method \cite{52} is applied to extract the edgemap of each natural image. During the testing phase, sketches, natural images, and the corresponding edgemaps are fed into the trained model, and the feature vectors of the three branches are obtained. Then, the cosine distance between the feature vectors of the query sketch and each natural image in the dataset is calculated. Finally, KNN is utilized to sort all the natural images for the final retrieval result. Similar to the existing SBIR methods, mean average precision (MAP) is used to evaluate the retrieval performance.

\subsection{Comparison Results}
Table \ref{table1} shows the performance of the proposed method compared with the state-of-the-art methods, on the Sketchy and TU-Berlin Extension datasets. The comparison methods include traditional methods (\emph{i.e.,} LKS \cite{13}, HOG \cite{20}, SHELO \cite{12}, and GF-HOG \cite{11}) and deep learning methods (\emph{i.e.,} Siamese CNN \cite{33}, SaN \cite{37}, GN Triplet \cite{38}, 3D shape \cite{36}, Siamese-AlexNet \cite{34}, Triplet-AlexNet \cite{34}, DSH \cite{34} and GDH \cite{39}). It can be observed from the table that: 1) Compared with the methods that utilize edgemaps to replace natural images for retrieval \cite{33}, \cite{36}, the proposed method obtains better retrieval accuracy. This is mainly because the edgemaps extracted from natural images may lose certain information useful for retrieval, and the CNNs pre-trained on ImageNet are more effective for natural images than edgemaps. 2) Compared with the methods that only utilize sketch-image pairs for SBIR \cite{38}, the proposed Semi3-Net achieves better performance, by introducing edgemap information. It demonstrates that an edgemap can be utilized as a bridge to effectively narrow the distance between the natural image and sketch domains. 3) Compared with DSH \cite{34}, which fuses the natural image and edgemap features into one feature representation, the proposed Semi3-Net achieves 0.133 and 0.230 improvements in terms of MAP, on Sketchy and TU-Berlin Extension, respectively. This is mainly because the proposed Semi3-Net makes full use of the one-to-one matching relationship between a natural image and its corresponding edgemap to align different domains into a high-level semantic space. Meanwhile, the domain shift process is well achieved under the proposed semi-heterogeneous joint embedding network architecture. Besides, by integrating the co-attention model and hybrid-loss mechanism, the proposed Semi3-Net is encouraged to learn a more discriminative embedding and invariant cross-domain representations, simultaneously. 4) The proposed Semi3-Net not only achieves superior performance over all traditional methods, but also outperforms the current best state-of-the-art deep learning method GDH \cite{39} by 0.106 and 0.110 in MAP on Sketchy and TU-Berlin Extension, respectively. This further validates the effectiveness of the proposed Semi3-Net for the SBIR task.
\begin{table}[!t]
\newcommand{\tabincell}[2]{\begin{tabular}{@{}#1@{}}#2\end{tabular}}
\renewcommand\arraystretch{1.25}
\centering
\footnotesize
\caption{\label{table1}COMPARISON WITH THE STATE-OF-THE-ART SBIR METHODS ON SKETCHY AND TU-BERLIN EXTENSION DATASETS.}
\begin{tabular}{p{3.5cm}<{\centering}|p{2cm}<{\centering}|p{2cm}<{\centering}}
\toprule[1.2pt]
\multirow{2}{*}{Methods}
& Sketchy & \tabincell{c}{TU-Berlin\\Extension}\\
\cline{2-3}
& MAP & MAP\\
\hline
 3D shape \cite{36} & $0.084$ & $0.054$ \\
HOG \cite{20} & $0.115$ & $0.091$ \\
GF-HOG \cite{11} & $0.157$ & $0.119$ \\
SHELO \cite{12} & $0.161$ & $0.123$ \\
 LKS \cite{13} & $0.190$ & $0.157$ \\
 SaN \cite{37} & $0.208$ & $0.154$ \\
 Siamese CNN \cite{33} & $0.481$ & $0.322$ \\
 Siamese-AlexNet \cite{34} & $0.518$ & $0.367$ \\
 GN Triplet \cite{38} & $0.529$ & $0.187$ \\
 Triplet-AlexNet \cite{34} & $0.573$ & $0.448$ \\
 DSH \cite{34} & $0.783$ & $0.570$ \\
GDH \cite{39} & $0.810$ & $0.690$ \\
Our method & $\bm{0.916}$ & $\bm{0.800}$ \\
\bottomrule[1.2pt]
\end{tabular}
\end{table}

\begin{figure*}[htbp]
\centering
\includegraphics[width=1\linewidth]{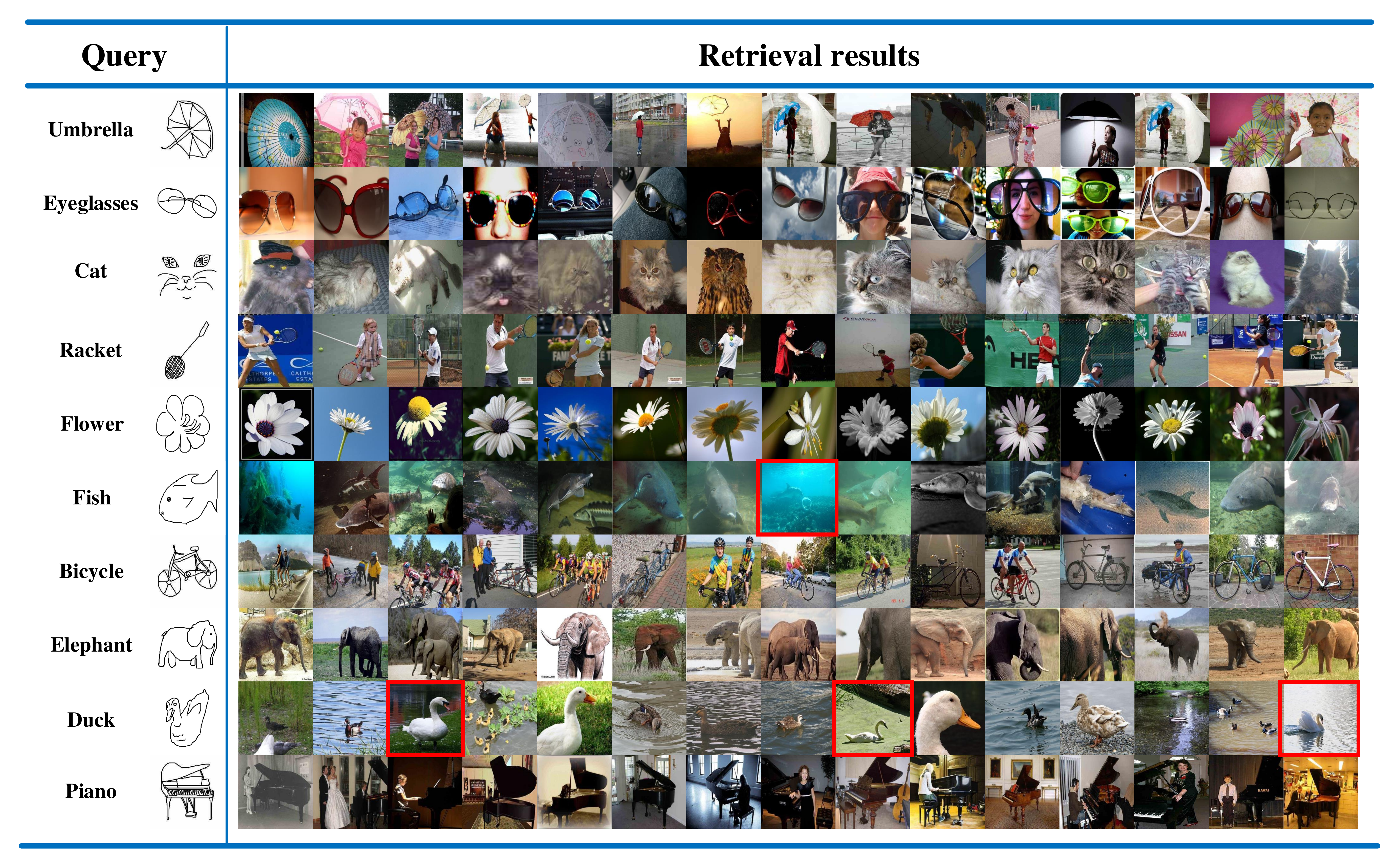}
\setlength{\abovecaptionskip}{-0.5cm}
\caption{ Some retrieval examples for the proposed Semi3-Net on the Sketchy dataset. Incorrect retrieval results are marked with red bounding boxes.}
\setlength{\belowcaptionskip}{-2cm}
\label{figure3}
\end{figure*}

Figure \ref{figure3} shows some retrieval examples by the proposed Semi3-Net for the Sketchy dataset. Specifically, 10 relatively challenging query sketches with top-15 retrieval rank lists are presented, where the incorrect retrieval results are marked with red bounding boxes. As can be seen, the proposed method performs well for retrieving natural images with complex backgrounds, such as the queries ``Umbrella'' and ``Racket''. This is mainly because the key information common to different domains can be effectively captured by the proposed semi-heterogeneous feature mapping part with the co-attention model. Additionally, for relatively abstract sketches, such as the sketch ``Cat'', the proposed method still achieves a consistent superior performance. This indicates that the semantic features of the abstract sketches are extracted by embedding the features from different domains into a high-level semantic space. In other words, although the abstract sketches only consist of black and white pixels, the proposed Semi3-Net can gain a better understanding of the sketch domain by introducing the joint semantic embedding part. However, there are also some incorrect retrieval images in Fig. \ref{figure3}. For example, for the query ``Fish'', a dolphin is retrieved with the proposed method. Meanwhile, for the query ``Duck'', several swans are retrieved. However, the incorrect retrieved natural images are quite similar to the query. Specifically, the fish and duck sketches, which lack color and texture information, are very similar in shape to a dolphin and a swan, respectively, which makes the SBIR task more challenging.

\begin{figure*}[htbp]
\centering
\includegraphics[width=1\linewidth]{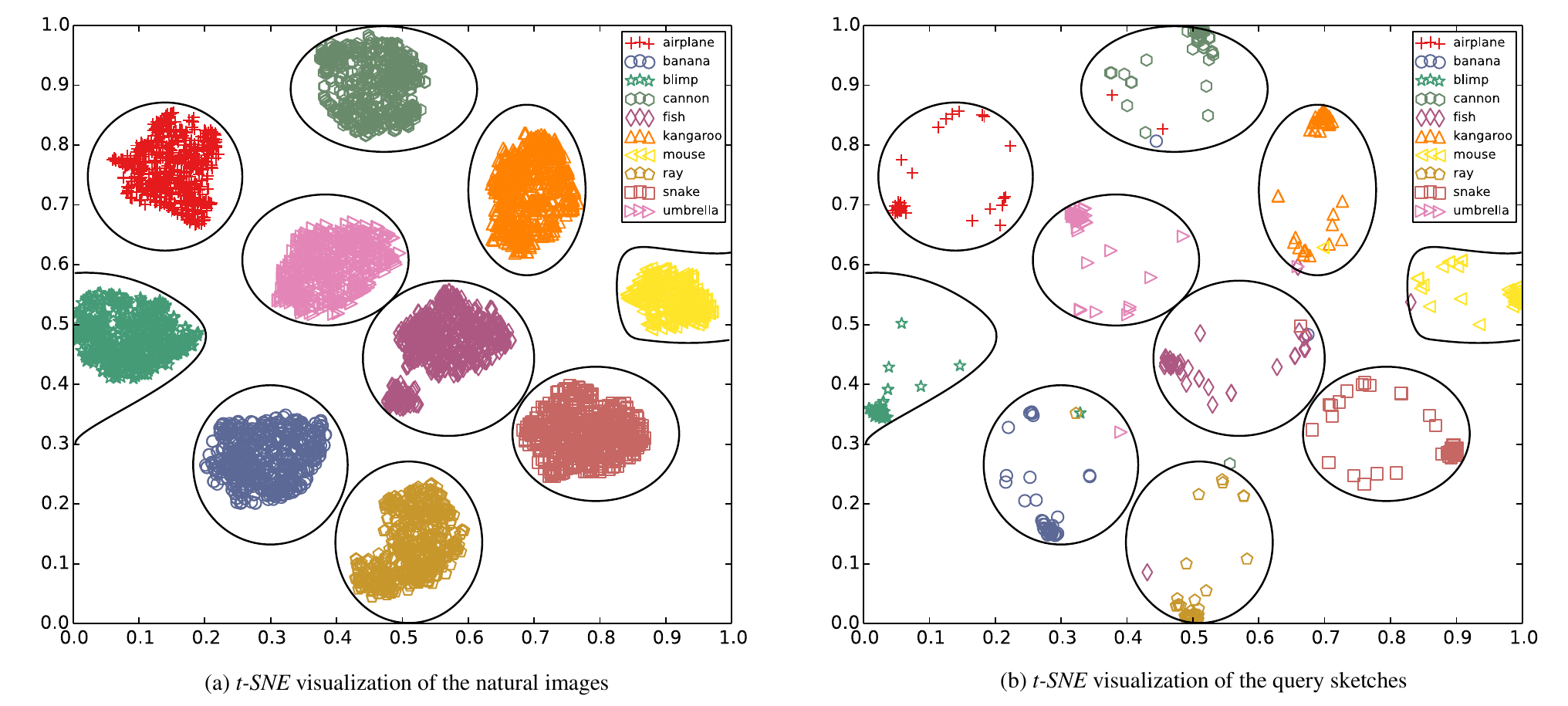}
\setlength{\abovecaptionskip}{-0.5cm}
\caption{\emph{t-SNE} visualization of the natural images and query sketches from ten categories in the Sketchy dataset, during the test phase. Symbols with the same color and shape in (a) and (b) represent the natural images and query sketches with the same label, respectively.}
\setlength{\belowcaptionskip}{-2cm}
\label{figure4}
\end{figure*}

\begin{table}[!t]
\renewcommand\arraystretch{1.2}
\centering
\footnotesize
\caption{\label{table2}EVALUATION OF THE SEMI-HETEROGENEOUS ARCHITECTURE ON SKETCHY DATASET.}
\begin{tabular}{p{5cm}<{\centering}|p{3cm}<{\centering}}
\toprule[1.2pt]
Methods & MAP \\
\hline
All sharing & $0.857$\\
Only FC layer sharing & $0.879$ \\
Only sketch-edgemap sharing & $0.890$ \\
Semi3-Net & $\bm{0.916}$ \\
\bottomrule[1.2pt]
\end{tabular}
\end{table}
To further verify the effectiveness of the proposed Semi3-Net, the \emph{t-SNE} \cite{53} visualization of natural images and sketches from ten categories on Sketchy are reported. As shown in Fig. \ref{figure4}, the ten categories are selected randomly in the dataset, and the labels of the selected categories are also illustrated in the upper right corner. We run \emph{t-SNE} visualization on both images and sketches together, then separate the projected data points to Fig. \ref{figure4} (a) and Fig. \ref{figure4} (b), respectively. Specifically, the circles in Fig. 4 represent clusters of different categories, and the data in each circle belongs to the same category in the high-level semantic space. In other words, circles in the same position of Fig. \ref{figure4} (a) and Fig. \ref{figure4} (b) correspond to the natural images and sketches with the same label, respectively. If data samples with the same label but from two different domains are correctly aligned in the common feature space, it indicates that the cross-domain learning is successful. As can be seen, the natural images and sketches with the same label scatter into nearly the same clusters. This further demonstrates that the proposed Semi3-Net effectively aligns sketches and natural images into a common high-level semantic space, thus improving the retrieval accuracy.

\subsection{Ablation Study}
\subsubsection{Evaluation of the semi-heterogeneous architecture}
To evaluate the superiority of the proposed semi-heterogeneous network architecture, networks with different weight sharing strategies are compared with the proposed Semi3-Net. For fair comparison, the proposed network architecture is fixed when different weight-sharing strategies are verified. In other words, both the proposed co-attention model and the hybrid-loss mechanism are applied in the networks with different weight-sharing strategies. The comparison results are shown in Table \ref{table2}, where ``all sharing'' indicates the weights of the three branches are completely shared, ``only FC layer sharing'' indicates only the weights of the three branches are shared in the semantic embedding part, and ``only sketch-edgemap sharing'' indicates only the weights of the sketch branch and edgemap branch are shared in the feature mapping part. As can be seen, the method with all sharing strategy does not perform as well as the methods with partially-shared strategies. This supports our views illustrated  previously that the bottom discriminative features should be learned separately for different domains. Compared with the the proposed Semi3-Net, the method with only FC layer sharing strategy obtains 0.879 MAP. That is because the intrinsic relevance between the sketches and edgemaps is ignored. Meanwhile, experimental results show that the method with only sketch-edgemap sharing strategy also results in a decrease in MAP. That is because the three branches are not fully aligned in the common high-level semantic space. Importantly, the proposed Semi3-Net, with both the FC layer sharing and sketch-edgemap sharing strategies, achieves the best performance, which proves the effectiveness of the proposed network architecture.

\subsubsection{Evaluation of key components}

In order to illustrate the contributions of the key components in the proposed method, \emph{i.e.,} the self-heterogeneous three-way framework, the co-attention model and the hybrid-loss mechanism, leaving-one-out evaluations are conducted on the Sketchy dataset. The experimental results are shown in Table \ref{table3}. Note that ``w/o CAM and w/o HLM'' refers to the proposed self-heterogeneous three-way framework, which has neither a co-attention model nor a hybrid-loss mechanism. For ``w/o HLM'', only two types of typical SBIR losses, the cross-entropy loss and the sketch-image contrastive loss, are used in the training phase. As shown in Table \ref{table3}, the proposed self-heterogeneous three-way framework obtains 0.851 MAP, outperforming the other methods in Table \ref{table1}. Additionally, leaving out either CAM or HLM results in a lower MAP, which verifies that each component contributes to the overall performance. Specifically, the informative features common between natural images and the corresponding edgemaps are captured effectively by the proposed co-attention mechanism, and the three different inputs can be effectively aligned into a common feature space by introducing the hybrid-loss mechanism.

\subsection{Discussions}

\begin{table}[!t]
\renewcommand\arraystretch{1.2}
\centering
\footnotesize
\caption{\label{table3}EVALUATION OF KEY COMPONENTS ON SKETCHY DATASET.}
\begin{tabular}{p{5cm}<{\centering}|p{3cm}<{\centering}}
\toprule[1.2pt]
Methods & MAP \\
\hline
w/o CAM and w/o HLM & $0.851$\\
w/o HLM & $0.880$ \\
w/o CAM & $0.896$ \\
Semi3-Net & $\bm{0.916}$ \\
\bottomrule[1.2pt]
\end{tabular}
\end{table}
\begin{table}[!t]
\renewcommand\arraystretch{1.3}
\centering
\footnotesize
\caption{\label{table4}EVALUATION OF DIFFERENT FEATURE SELECTIONS IN THE TEST PHASE ON SKETCHY DATASET.}
\begin{tabular}{p{5cm}<{\centering}|p{3cm}<{\centering}}
\toprule[1.2pt]
Methods & MAP \\
\hline
Edgemap feature & $0.914$\\
Natural image feature & $\bm{0.916}$ \\
\bottomrule[1.2pt]
\end{tabular}
\end{table}
In this section, the impact of different feature selections in the test phase is discussed for the Sketchy dataset. As mentioned above, either features extracted from the natural image branch or the edgemap branch can be used as the final retrieval feature representation. To evaluate the impact of different feature selections in the test phase, we obtain the 256-d feature vectors from the embedding layers of the natural image branch and the edgemap branch, respectively. The experiments are conducted on the Sketchy dataset, and the experimental results are reported in Table \ref{table4}. As can be seen from the table, there is a small difference between the retrieval performances when using the edgemap feature and natural image feature. This is mainly because the invariant cross-domain representations of different domains are indeed learned by the proposed method, and the cross-domain gap is narrowed by the joint embedding learning. Besides, the feature extracted from the natural image branch performs a little better than that from the edgemap branch, which is also consistent with the research in \cite{30}, \cite{38}.

In addition, to verify the performance without edgemap branch, we conducted experiments on a two-branch framework by only using the sketch and natural image branches. Considering that the sketches and natural images belong to two different domains, a non-shared setting is first exploited on the two-branch framework. Note that without the edgemap branch, neither the co-attention model nor the hybrid-loss mechanism is added into the two-branch framework. Compared with the proposed Semi3-Net with 0.916 MAP, the two-branch framework obtains 0.837 MAP on Sketchy dataset, which sufficiently proves the effectiveness of the edgemap branch in the proposed Semi3-Net. In addition, a partially-shared setting is also exploited on the two-branch framework, in which the weights of the convolutional layers are independent while the weights of the fully-connected layers are shared. Compared with the non-shared setting, the partially-shared setting obtains 0.861 MAP. This further verifies the importance of the joint semantic embedding for SBIR.

\section{Conclusion}
In this paper, we propose a novel semi-heterogeneous three-way joint embedding network, where auxiliary edgemap information is introduced as a bridge to narrow the cross-domain gap between sketches and natural images. The semi-heterogeneous feature mapping and the joint semantic embedding are proposed to learn the specific bottom features from each domain and embed different domains into a common high-level semantic space, respectively. Besides, a co-attention model is proposed to capture the informative features common between natural images and the corresponding edgemaps, by recalibrating corresponding channel-wise feature responses. In addition, a hybrid-loss mechanism is designed to construct the correlation among sketches, edgemaps, and natural images, so that the invariant cross-domain representations of different domains can be effectively learned. Experimental results on two datasets demonstrate that Semi3-Net outperforms the state-of-the-art methods, which proves the effectiveness of the proposed method.

In the future, we will focus on extending the proposed cross-domain network to fine-grained image retrieval, and learning the correspondence of the fine-grained details for sketch-image pairs. Besides, further study may also include extending our method to other cross-domain learning problems.

\ifCLASSOPTIONcaptionsoff
  \newpage
\fi

\begin{IEEEbiography}[{\includegraphics[width=1in,height=1.25in,clip,keepaspectratio]{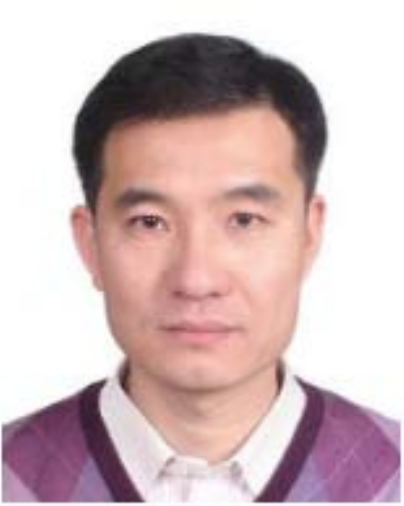}}]{Jianjun Lei}
(M'11-SM'17) received the Ph.D. degree in signal and information processing from Beijing University of Posts and Telecommunications, Beijing, China, in 2007.
He was a visiting researcher at the Department of Electrical Engineering, University of Washington, Seattle, WA, from August 2012 to August 2013. He is currently a Professor at Tianjin University, Tianjin, China. His research interests include 3D video processing, virtual reality, and artificial intelligence.
\end{IEEEbiography}

\begin{IEEEbiography}[{\includegraphics[width=1in,height=1.25in,clip,keepaspectratio]{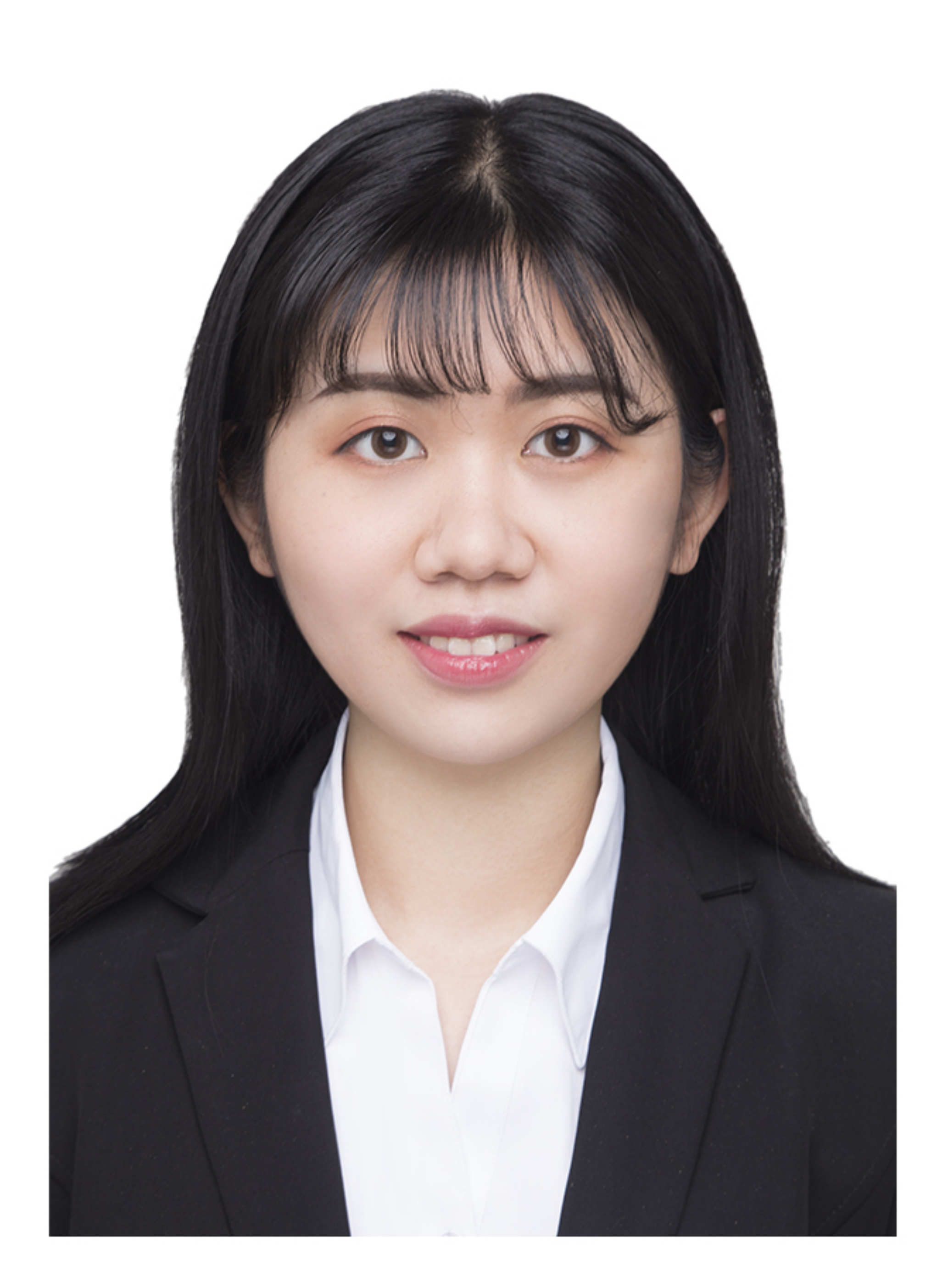}}]{Yuxin Song}
received the B.S. degree in communication engineering from Hefei University of Technology, Hefei, Anhui, China, in 2017. She is currently pursuing the M.S. degree with the School of Electrical and Information Engineering, Tianjin University, Tianjin, China. Her research interests include image retrieval and deep learning.
\end{IEEEbiography}

\begin{IEEEbiography}[{\includegraphics[width=1in,height=1.25in,clip,keepaspectratio]{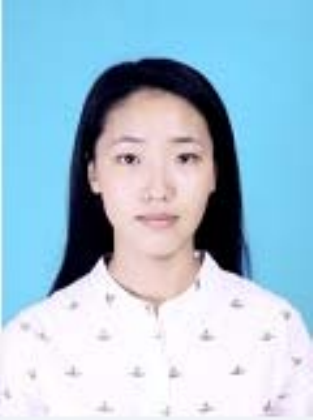}}]{Bo Peng}
received the M.S. degree in communication and information systems from Xidian University, Xi'an, Shaanxi, China, in 2016. Currently, she is pursuing the Ph.D. degree at the School of Electrical and Information Engineering, Tianjin University, Tianjin, China. Her research interests include computer vision, image processing and video action analysis.
\end{IEEEbiography}

\begin{IEEEbiography}[{\includegraphics[width=1in,height=1.25in,clip,keepaspectratio]{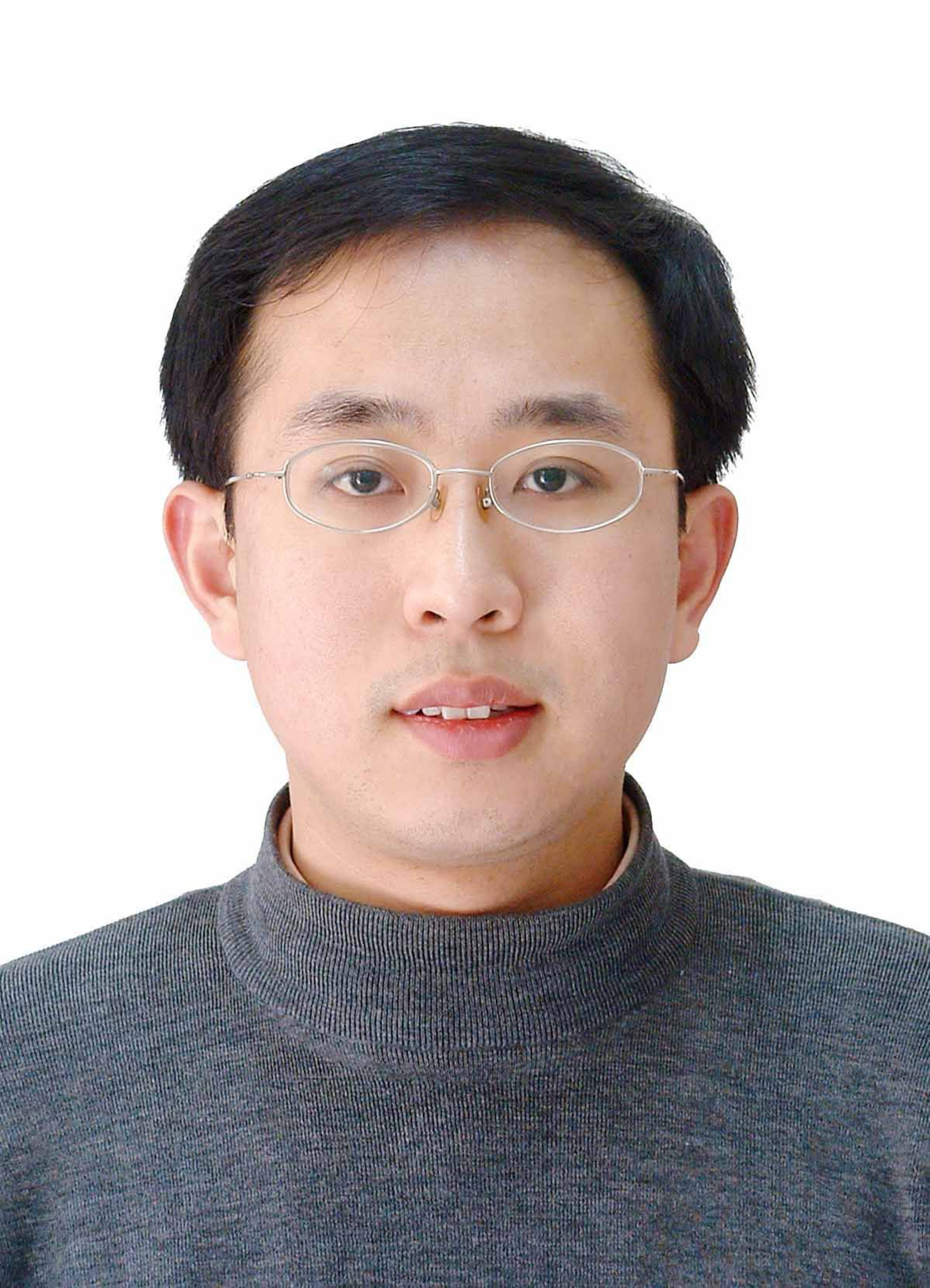}}]{Zhanyu Ma}
has been an Associate Professor at Beijing University of Posts and Telecommunications, Beijing, China, since 2014. He is also an adjunct Associate Professor at Aalborg University, Aalborg, Denmark, since 2015. He received his Ph.D. degree in Electrical Engineering from KTH (Royal Institute of Technology), Sweden, in 2011. From 2012 to 2013, he has been a Postdoctoral research fellow in the School of Electrical Engineering, KTH, Sweden. His research interests include pattern recognition and machine learning fundamentals with a focus on applications in computer vision, multimedia signal processing, and data mining.
\end{IEEEbiography}

\begin{IEEEbiography}[{\includegraphics[width=1in,height=1.25in,clip,keepaspectratio]{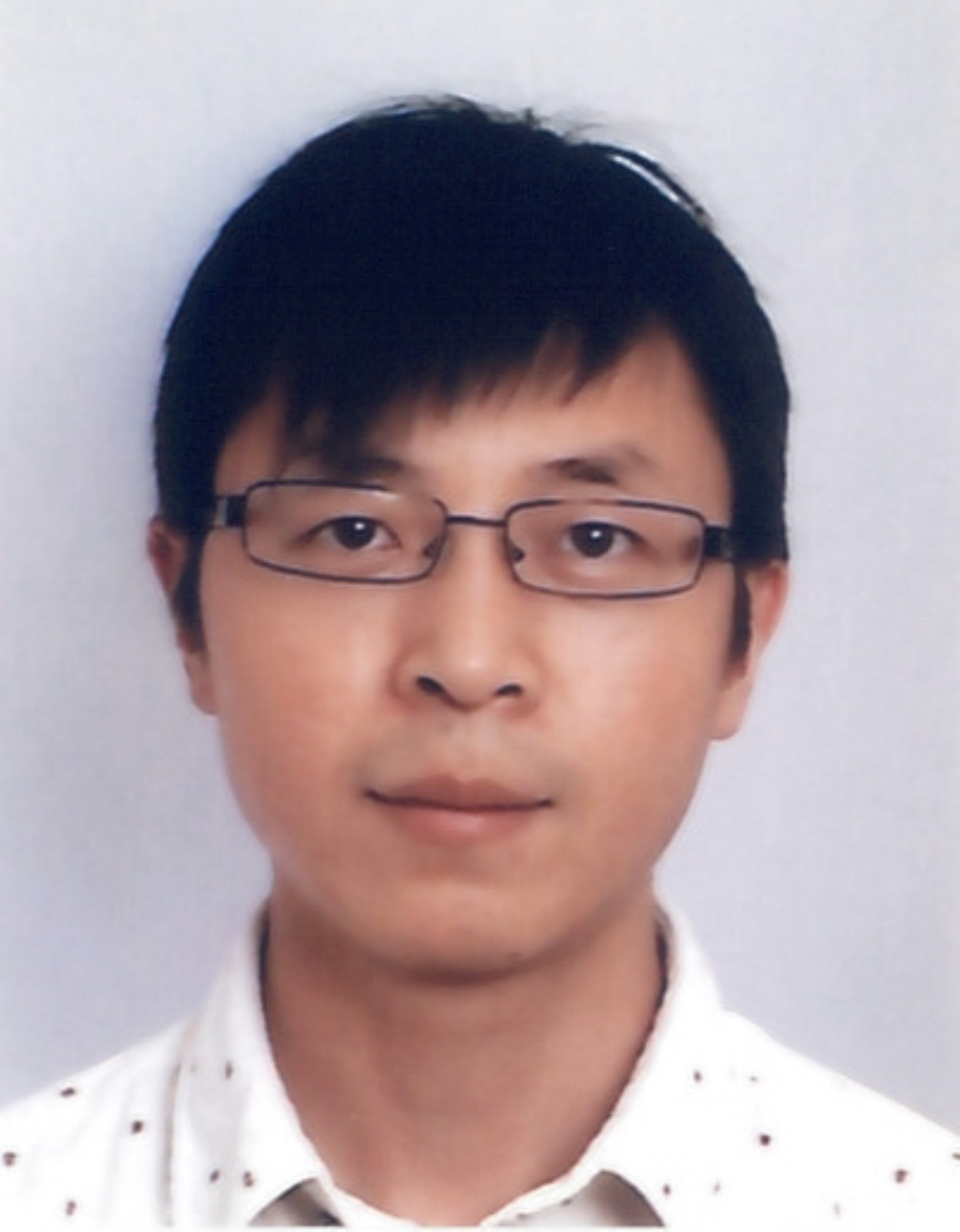}}]{Ling Shao}
is the CEO and Chief Scientist of the Inception Institute of Artificial Intelligence, Abu Dhabi, United Arab Emirates. His research interests include computer vision, machine learning, and medical imaging. He is a fellow of IAPR, IET, and BCS. He is an Associate Editor of the IEEE Transactions on Neural Networks and Learning Systems, and several other journals.
\end{IEEEbiography}

\begin{IEEEbiography}[{\includegraphics[width=1in,height=1.25in,clip,keepaspectratio]{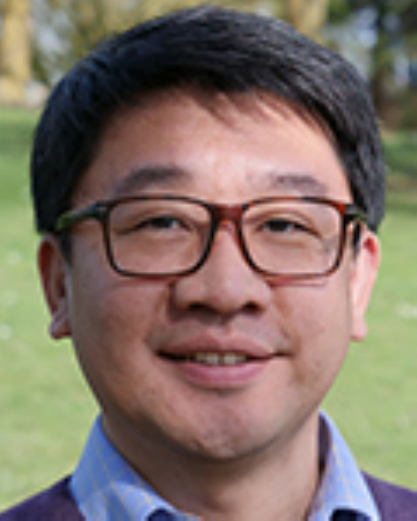}}]{Yi-Zhe Song}
is a Reader of Computer Vision and Machine Learning at the Centre for Vision Speech and Signal Processing (CVSSP), UK¡¯s largest academic research centre for Artificial Intelligence with approx. 200 researchers. Previously, he was a Senior Lecturer at the Queen Mary University of London, and a Research and Teaching Fellow at the University of Bath. He obtained his PhD in 2008 on Computer Vision and Machine Learning from the University of Bath, and received a Best Dissertation Award from his MSc degree at the University of Cambridge in 2004, after getting a First Class Honours degree from the University of Bath in 2003. He is a Senior Member of IEEE, and a Fellow of the Higher Education Academy. He is a full member of the review college of the Engineering and Physical Sciences Research Council (EPSRC), the UK's main agency for funding research in engineering and the physical sciences, and serves as an expert reviewer for the Czech National Science Foundation.
\end{IEEEbiography}

\end{document}